\documentclass[conference]{IEEEtran}
\usepackage{cite}
\usepackage{url}
\usepackage{amsmath,amssymb,amsfonts}
\usepackage{algorithmic}
\usepackage{graphicx}
\usepackage{textcomp}
\usepackage{xcolor}
\usepackage[numbers]{natbib}
\def\BibTeX{{\rm B\kern-.05em{\sc i\kern-.025em b}\kern-.08em
    T\kern-.1667em\lower.7ex\hbox{E}\kern-.125emX}}
\begin{document}

\title{Assessing Guest Nationality Composition\\ from Hotel Reviews}

\author{\IEEEauthorblockN{Fabian Gröger}
    \IEEEauthorblockA{\small
    	\textit{Lucerne University of} \\ \textit{Applied Sciences and Arts}\\
    	Rotkreuz, Switzerland \\
    	fabian.groeger@hslu.ch}
\and
\IEEEauthorblockN{Marc Pouly}
  \IEEEauthorblockA{\small
    	\textit{Lucerne University of} \\ \textit{Applied Sciences and Arts}\\
    	Rotkreuz, Switzerland \\
    	marc.pouly@hslu.ch}
\and
\IEEEauthorblockN{Flavia Tinner}
    \IEEEauthorblockA{\small
    	\textit{University of Lucerne} \\
    	Lucerne, Switzerland \\
    	flavia.tinner@unilu.ch}
\and
\IEEEauthorblockN{Leif Brandes}
    \IEEEauthorblockA{\small
    	\textit{University of Lucerne} \\
    	Lucerne, Switzerland \\
    	leif.brandes@unilu.ch}
}

\maketitle

\begin{abstract}
Many hotels target guest acquisition efforts to specific markets in order to best anticipate individual preferences and needs of their guests.\ Likewise, such strategic positioning is a prerequisite for efficient marketing budget allocation.\ Official statistics report on the number of visitors from different countries, but no fine-grained information on the guest composition of individual businesses exists.\ There is, however, growing interest in such data from competitors, suppliers, researchers and the general public.\ We demonstrate how machine learning can be leveraged to extract references to guest nationalities from unstructured text reviews in order to dynamically assess and monitor the dynamics of guest composition of individual businesses.\ In particular, we show that a rather simple architecture of pre-trained embeddings and stacked LSTM layers provides a better performance-runtime tradeoff than more complex state-of-the-art language models.
\end{abstract}


\section{Introduction}

Hardly any other industry was hit as much by the pandemic as the tourism industry.\ In 2020, and according to the World Tourism Organization, the number of overnight stays in the European Union dropped by $1.1$ billion or $57\%$.\ In Switzerland, there was a milder but still considerable loss of $40\%$ in the same year.\ International travel restrictions and change in guest behavior were identified as the driving factors.\ In fact, in the ten years leading up to the Covid-19 outbreak, we can observe a clear focus shift away from local and nearby markets towards distant and emerging markets.\ For example, Switzerland reports an increase in the number of overnight stays from Chinese travelers by $412\%$ during this period; there were $144\%$ more bookings from Indian guests and $79\%$ more from US citizens.\ At the same time, overnight stays from European guests dropped by $28\%$.\ In 2019, $55\%$ of all overnight stays in Switzerland were booked by foreign guests; $24\%$ came from outside Europe \cite{HotellerieSuisse.2021}. 

With ever increasing demand from international guests, many hotels and resorts were seen to position themselves more strategically and concentrate guest acquisition efforts to specific markets, nationalities and culture groups.\ The reasons to specialize are manyfold:\ guests with different cultural background have different preferences, culinary restrictions and infrastructural needs (e.g., halal-friendly hotels).\ They appreciate front-desk staff with dedicated language and inter-cultural competences, etc.\ Clear target markets are further required for efficient allocation of marketing budget or the evaluation of strategic partnerships.\ 

% use different booking and review platforms and expect them to be managed actively

Statistical authorities\footnote{Switzerland:\ Beherbergungsstatistik HESTA, Bundesamt für Statistik} and tourism associations regularly report on guest composition per country or greater region.\ These figures provide important evidence for risk assessment, political discussion and decision making.\ For other purposes, however, more fine-grained information is necessary.\ For example, hotels and resorts can optimize their strategic positioning with respect to the main acquisition markets of their competitors.\ Likewise, they can pursue a different dynamic pricing strategy knowing what markets they share to which extend with their competitors.\ Suppliers of goods and services to hotels, on the other hand, can extend their offering to the specific needs of travel groups.\ And restaurants can dynamically react to culinary preferences of visitors hosted by the nearby hotels.\ Finally, there is a strong interest from researchers and the general public in this data as it would provide evidence on cluster risks, over-tourism, etc.\ down to municipalities.

However, gathering data about the guest composition of individual businesses (hotels, resorts, districts, etc.) is currently unsolved.\ Businesses are unlikely to reveal this data.\ They consider it confidential, do not want to publicly expose or commit to a guest acquisition strategy.\ Also, guest composition changes over time and is subject to external influences such as political conflicts or economic recession in source countries.\ It therefore cannot be considered static data but needs to be monitored dynamically over time.\ Finally, manual assessment will clearly never scale up to an interactive map of guest compositions per business in a larger city or tourist destination. 

\section{Nationality Detection in Reviews}

Online booking platforms often display the nationality of registered reviewers by a flag icon.\ But sometimes, the flag only indicates the IP range from where the review was written.\ Also, not all nationalities are equally keen on writing reviews, which can potentially introduce a strong selection bias.\ However, a surprisingly large number of reviews explicitly mention the nationality of other guests, and we consider this information to be less biased.\ Hence, instead of focussing on the nationality of the review authors, we suggest a natural language processing pipeline to extract guest nationality references from unstructured reviews. 

Even though the number of official German nationality terms is fixed, a simple filtering approach based on a public dictionary does not work.\ Dictionaries do not include slang or insulting terms like \emph{Amis} for \emph{Amerikaner}, and many nationality terms actually do not refer to other guests.\ A typical example is:\ \emph{Das Essen beim Italiener was ausgezeichnet}.\ We use an extended filtering approach only to ensure that training and test data contain enough nationality terms and use a supervised learning approach for guest nationality detection even in the presence of out-of-vocabulary terms. 

\section{Datasets and Annotation}

A private dataset of $1.1\text{M}$ German hotel reviews (only confirmed bookings) was made available to us by a large booking platform.\ We used these reviews for training and hyper-parameter optimization.\ The reviews were split into $2.7\text{M}$ unique sentences.\ We used a public dictionary of $203$ nationality terms\footnote{\label{lingolia}\url{https://deutsch.lingolia.com/de/wortschatz/laender-nationalitaeten}, accessed on 23.11.2021} for pre-filtering enriched with the 10 nearest neighbors of each term from a pre-trained FastText model to also incorporate frequently used slang terms.\ $680k$ sentences contain at least one vocabulary term.\ We randomly sampled and annotated $750$ sentences from both subsets and split into $70\%$ training and $30\%$ validation set.

Test data, on the other hand, was taken from an open-source German review dataset \cite{Guhr.2020.Sentiment}, crawled from \url{holidaycheck.de} and consisting of more than $3.6 \text{M}$ anonymous reviews.\ The data was pre-processed and filtered identically, and $75$ sentences were randomly sampled with and without nationality terms.\ Three external annotators were provided with a definition of guest nationality reference and some examples (e.g., the Italian restaurant sentence as a negative example).\ The annotators achieved a Fleiss kappa score of $91.2\%$ showing almost perfect agreement. 

\section{Results}

Multiple machine learning models of varying complexity were trained on the training set.\ The results from the validation set are displayed in Table \ref{tab1}.

\begin{table}
    \centering
    \caption{Model comparison on validation set.}
    \label{tab1}
    \begin{tabular}{l | l l l l }%
    \hline
        \bfseries Model & \bfseries Pr & \bfseries Re & \bfseries Acc & \bfseries $\text{F}_1$\\
    \hline
        Dictionary based & 40.9 & \textbf{100.0} & 58.1 & 40.9 \\ 
    \hline
        TF-IDF + SVC & 85.1 & 75.3 & 79.1 & 76.1 \\
    \hline
        Embedding Layer + LSTM & 95.2 & 95.4 & 94.3 & 94.7 \\
        Embedding Layer + BiLSTM & 97.6 & 93.8 & 95.5 & 95.8 \\
        FastText + LSTM & 97.6 & 93.0 & 96.6 & 96.2 \\
        FastText + BiLSTM & \textbf{99.0}  & 99.0 & \textbf{98.9} & \textbf{99.0} \\
    \hline
        BERT & 97.9 & 96.9 & 97.7 & 97.4 \\
    \hline
    \end{tabular} \vspace{-.5cm}
\end{table}

The best performing model consists of pre-trained FastText embeddings followed by a stack of bidirectional LSTM layers.\ This rather simple architecture performs more that $20\%$ better in $\text{F}_1$ compared to a support vector machine baseline with TF-IDF feature engineering and also beats transfer learning with a pretrained BERT model\footnote{\url{www.huggingface.co/bert-base-german-cased}, accessed on 09.02.2022} by a small margin.\ On the one hand, this indicates that such large language models as BERT can generalize pretty well even in the presence of small data.\ On the other hand, BERT comes with an increase of $160\%$ in the number of parameters and thus significantly higher inference time compared to the winning model.\ Hence, even with more data and an expected increase in the BERT model performance, the LSTM architecture will show the better performance-runtime tradeoff. 

Table \ref{tab2} shows an excerpt of a qualitative evaluation of the best performing model by displaying its predictions in specific edge cases.\ On the hidden test set prepared with independent annotators, the winning model achieved an $\text{F}_1$ score of $93.2\%$, convincingly demonstrating its capability to generalize to reviews from other sources.\ For comparison, the keyword-based approach without supervised learning would only achieve an $\text{F}_1$ score as low as $31.3\%$.

%\section{Related Work}
% Hotel Domain
%Machine learning has been applied to the hotel industry to predict the number of booking cancellations \cite{b1, b2, b3}, to forecast the demand \cite{b4, b5}, to evaluate the hotel location \cite{b6}, to predict hotel maintenance \cite{b7}, to predict hotel energy demand \cite{b8} and to forecast hotel room prices \cite{b9}.
% Detecting Nationalities
%Inferring nationalities from texts has been studied in other works, such as detecting the nationality from a surname \cite{b10} or from Twitter usernames \cite{b11}. To the best of our knowledge, this is the first attempt to use the detected nationalities in reviews for helping to optimize the hotel industry. 
% TODO: Related work transformers
% Needed?

 \begin{table}
    \centering
    \caption{Qualitative analysis of the winning model.\ True indicates a detected guest nationality reference; it is written in boldface when the model prediction was correct.\ Typos (\emph{Afgahnen} and \emph{Andoranern}) were made intentionally to test out-of-vocabulary behavior. }
    \label{tab2}
    \begin{tabular}{l | c}
         \hline
         \textbf{Sentence} & \textbf{Prediction} \\
         \hline
         Das Hotel war komplett voll mit Andoranern. & \textbf{True} \\
         Beim Afgahnen war das Essen vorzüglich. & \textbf{False} \\
         Die Amis sind wieder negativ aufgefallen. & \textbf{True} \\ \hline
         Im Zentrum gibt es auch Pubs, Pizza, Chinesen etc. & \textbf{False} \\
         Beim Italiener war das Essen fantastisch. & \textbf{False} \\
         Richtung inland gibt es viele Italiener die preiswert sind. & True \\ \hline
    \end{tabular} \vspace{-.4cm}
\end{table}

\section{Conclusion and Future Work}

Various machine learning models were trained and evaluated on detecting references to guest nationalities in unstructured hotel reviews for the purpose of dynamically assessing and monitoring guest composition of individual businesses.\ A rather simple model of pre-trained embeddings and stacked LSTM layers achieved $93.2\%$ $\text{F}_1$ on unseen public test data.\ Moreover, it was found to offer a better performance-runtime tradeoff than large state-of-the-art language models for this particular task.\ A qualitative analysis demonstrates that the model can successfully deal with out-of-vocabulary terms.\ As future work, we plan to make available an interactive map overlay that displays the estimated guest composition of tourism businesses with sufficient reviews.

% Conclusion
%We trained a machine learning model to detect nationalities using a small dataset and achieved an $\text{F}_1$ score of 93.2\% on the test set. Evaluation showed that nationality detection is a problem that can be solved well with traditional algorithms and does not need much data to perform reasonably well and generalize to other corpora. 
% Future work
%In future work, our proposed nationality detection method can be used by the hotel industry to analyze their competitors' guest composition. Further, it can be used to build a co-occurrence matrix of nationalities which can then, in turn, be used to optimize the hotel's group-specific offers.

% Gästestruktur pro Konkurrenzbetrieb (hier würde ich es auch auf Regionsebene machen, aus Sicht eines Restaurants/ Retailers) als Google Maps Overlay
% Co-Occurence von Nationalitäten, etc. [hier habe ich mich noch gefragt, ob man changes abbilden kann, aber wohl nicht, wenn wir die 2018 Daten nehmen, oder?)

\bibliography{Citations}
\bibliographystyle{ieeetran}

\end{document}